\documentclass[9pt,draftcls,onecolumn]{IEEEtran}
\usepackage{graphicx}
\usepackage{subfigure}
\ifCLASSINFOpdf

\begin{document}
%
\title{A Fast Hierarchical Multilevel Image Segmentation Method using Unbiased Estimators}

\author{Sreechakra~Goparaju,~\IEEEmembership{Student~Member,~IEEE,}
        Jayadev~Acharya,
        Ajoy~K.~Ray,~\IEEEmembership{Member,~IEEE,}
        Jaideva~C.~Goswami,~\IEEEmembership{Senior~Member,~IEEE}%

\thanks{G. Sreechakra and A. K. Ray are with the Department of Electronics and Electrical Communication Engineering, Indian Institute of Technology, Kharagpur, WB - 721302, India. Email: sreechakra@gmail.com, akray@smst.iitkgp.ernet.in}
\thanks{J. Acharya is with the Department
of Electrical and Computer Engineering, University of California San Diego, La Jolla, CA - 92093, USA Email: jacharya@ucsd.edu .}
\thanks{J. C. Goswami is Principal Engineer at Schlumberger Technology Center, Sugar Land, TX - 77478, USA. He was a Professor at IIT Kharagpur. Email: jcgoswami@ieee.org}}%

\maketitle

\begin{abstract}
This paper proposes a novel method for segmentation of images by
hierarchical multilevel thresholding. The method is global,
agglomerative in nature and disregards pixel locations. It involves
the optimization of the ratio of the unbiased estimators of within
class to between class variances. We obtain a recursive relation at
each step for the variances which expedites the process. The
efficacy of the method is shown in a comparison with some well-known
methods.
\end{abstract}


\IEEEpeerreviewmaketitle

\section{Introduction}
Thresholding is an important technique for image segmentation and
object extraction. The aim of an effective segmentation is to
separate objects from the background and to differentiate pixels
having nearby values for improving the contrast. The use of binary
images reduces the computational cost of the succeeding steps and
increases the ease of manipulation compared to using gray-level
images. Image thresholding is a well-researched field and there
exist many algorithms for determining an optimal threshold of the
image. Many surveys of thresholding methods and their applications
exist in literature~\cite{chietal},~\cite{Sezgin}.

Thresholding techniques can be divided into bilevel and multilevel
categories, depending on the number of classes. In bilevel
thresholding, the image is segmented into two different regions. The
pixels with gray-level values greater than a certain value T are
classified as object pixels, and the others with values lesser than
T are classified as background pixels or vice versa. Otsu's
method~\cite{Otsu} chooses optimal thresholds by maximizing the
between class variance and minimizing within class variance. Sahoo
et al.~\cite{Sahoo} found that in global thresholding, Otsu's method
is one of the better threshold selection methods for general real
world images with regard to uniformity and shape measures. However,
inefficient formulation of between class variance makes the method
time consuming. Abutaleb~\cite{abutaleb} used two-dimensional
entropy to calculate the threshold. Kittler and
Illingworth~\cite{Kittler} suggested a minimum error thresholding
method. They assume that an image is characterized by a mixture
distribution with both object and background classes having a
Gaussian distribution. The probability density function for each
class is estimated based on the histogram of the image.
Kwon~\cite{kwon} proposed a threshold selection method based on
cluster analysis. The method proposed a criterion function that
involved not only the histogram of the image but also the
information on spatial distribution of pixels. This criterion
function optimizes intra-class similarity to achieve the most
similar class and inter-class similarity to confirm that every
cluster is well separated. The similarity function uses all pixels
in the two clusters as denoted by their coordinates.

Multilevel thresholding is a process that segments a gray-level
image into several distinct regions. The method works very well for
objects with colored or complex backgrounds, on which bilevel
thresholding fails to produce satisfactory results. Reddi et
al.~\cite{Reddi} proposed an iterative form of Otsu's method, so as
to generalize it to multilevel thresholding. Ridler and Calward
algorithm~\cite{Ridler} uses an iterative clustering approach. An
initial estimate of the threshold is made (e.g., mean image
intensity) and pixels above and below are assigned to the white and
black classes respectively. The threshold is then iteratively
re-estimated as the mean of two class means. The most difficult task
is to determine the appropriate number of thresholds automatically.
Papamarkos and Gatos~\cite{Papa} specify the multithreshold values
as the global minima of the rational functions which approximate the
histogram segments by using hill clustering technique to determine
the peak locations of image histogram. Arora et al.~\cite{Arora}
propose a simple multilevel thresholding using means and variances.
Here we also point out that the extension of various binarization
methods become extremely complex for multilevel segmentation due to
the curse of dimensionality. We will discuss this with respect to
Otsu's method. Arifin and Asano~\cite{Arifin} propose a hierarchical
clustering method, which is shown to be more efficient than some
existing methods. In their method, they use agglomerative clustering
(using a dendrogram of gray levels) based on a similarity measure
which involves the interclass variance of the clusters to be merged
and the intra-class variance of the newly merged cluster.

We use unbiased estimators of variances for optimization and find a
recursive formulation which improves computational efficiency of the
algorithm. We then compare our results with these well known global
thresholding methods and find that the proposed approach is better
than most existing techniques on the criteria that we consider.


\section{Approach and Algorithm}
In the proposed approach, we will optimize the ratio of unbiased
estimators of within and between class variances of the image to
find thresholds for segmenting the image into multiple levels. The
method is agglomerative hierarchical where we merge two classes at
each level, and replace the pixels belonging to either class by
their weighted mean gray-level. Thus we reduce the number of classes
by one at each iteration. This procedure can be easily imagined in
terms of a dendrogram. A similar approach has been used for
petrophysical data classification in~\cite{Moghaddamjoo} and by
Acharya et al. in~\cite{Acharya}. Let $G$ be the number of levels of
the gray scale image, so that the number of classes initially is
$G$. Also let $M$ be the \emph{a priori} known desired number of
classes. Let $t_{i}$ be the $i^{th}$ threshold at a given stage and
$C_{k}$ be the $k^{th}$ class, defined as $\{\hspace{1 mm} g \mid
t_{k-1} < g \leq t_{k}\}$, where $g$ denotes a gray-level between
$1$ and $G$. The thresholds $t_{0}$ and $t_{M}$ are defined to be 0
and $G$, respectively.

A histogram $h(g)$ is the function which shows the number of
occurrences of the gray-level $g$ in the image. If $N$ is the total
number of pixels in the image then, $N$ = $\sum_{g=1}^{G}h(g)$

$K(i)$ = the number of classes after the $i^{th}$ recursion of the
algorithm. Thus $K(0)=G$.

$n_{k}(i)$ = the number of pixels in the $k^{th}$ class after the
$i^{th}$ recursion

$a_{k}(i)$ = the amplitude of the $k^{th}$ class after $i^{th}$
recursion,

$a$ = the grand weighted mean gray-level of all the pixels in the
image. Note that $a_{k}(i)$ refers to the weighted pixel value (or
gray-level) of the pixels in class $C_{k}(i)$, i.e., class $C_{k}$
at the $i^{th}$ recursion.

$x_{km}(i)$ = the $m^{th}$ pixel in the $k^{th}$ class after the
$i^{th}$ recursion. Note that wherever $i$ is not defined, it refers
to the recursion number.

We define the unbiased estimator of within-class variance $v(i)$ and
that of between-class variance $w(i)$ at the $i^{th}$ recursion by
the following equations:
\begin{equation}
\label{v} v(i) = \frac{1}{N-K(i)}
\sum_{k=1}^{K(i)}\sum_{m=1}^{n_{k}(i)} [x_{km}(i)-a_{k}(i)]^2
\end{equation}
\begin{equation}
\label{w} w(i) = \frac{1}{K(i)-1} \sum_{k=1}^{K(i)}
n_{k}(i)[a_{k}(i)-a]^2
\end{equation}

The criterion function is given by $q(i)$
\begin{equation}
\label{q} q(i) = \frac{v(i)}{w(i)}
\end{equation}
where $q(i)$ is the value of the function to be minimized after the
$i^{th}$ iteration.

An exhaustive search which calculates the objective function for
every pair of connected classes eligible to be merged is
computationally cumbersome. We derive a recursive formulation which avoids this. Suppose we
are in the $(i-1)^{th}$ stage and the pair of classes to be combined are
$C_{l}(i-1)$ and $C_{l+1}(i-1)$. From equation \ref{v}, we have
\begin{eqnarray*}
v(i-1) &=& \frac{1}{N-K(i-1)}
\sum_{k=1}^{K(i-1)}\sum_{m=1}^{n_{k}(i-1)}
[x_{km}(i-1)-a_{k}(i-1)]^2 \\ &=& \frac{1}{N-K(i-1)}\Big\{\sum_{k=1,
k\neq l,
l+1}^{K(i-1)}\Big(\sum_{m=1}^{n_{k}(i-1)}[x_{km}(i-1)-a_{k}(i-1)]^2\Big)
\\ && +\sum_{m=1}^{n_{l}(i-1)} [x_{lm}(i-1)-a_{l}(i-1)]^2 \\ &&
+\sum_{m=1}^{n_{l+1}(i-1)} [x_{l+1,m}(i-1)-a_{l+1}(i-1)]^2\Big\}
\end{eqnarray*}

After $C_{l}(i-1)$ and $C_{l+1}(i-1)$ are merged, the number of
classes will be reduced by one ($K(i) = K(i-1)-1$) and the last two
terms above will be changed. Writing $v(i)$ in terms of the
parameters in the $(i-1)^{th}$ recursion, we get

\begin{eqnarray*}
v(i) &=& \frac{1}{N-K(i-1)+1}\Big\{\sum_{k=1, k\neq l,
l+1}^{K(i-1)}\Big(\sum_{m=1}^{n_{k}(i-1)}[x_{km}(i-1)-a_{k}(i-1)]^2\Big)
\\ && +\sum_{m=1}^{n_{l}(i-1)} [x_{lm}(i-1)-a_{l'}(i)]^2
+\sum_{m=1}^{n_{l+1}(i-1)} [x_{l+1,m}(i-1)-a_{l'}(i)]^2\Big\}
\end{eqnarray*}

where,

\begin{equation}
\label{a'} a_{l'}(i) = \frac
{n_{l}(i-1)a_{l}(i-1)+n_{l+1}(i-1)a_{l+1}(i-1)}{n_{l}(i-1)+n_{l+1}(i-1)}
\end{equation}

is the combined mean gray-level of the classes $C_{l}(i-1)$ and
$C_{l+1}(i-1)$ after they merge.

Comparing the above few equations, and doing some algebraic
manipulations, we ultimately get

\begin{equation}
\label{vv} v(i) = \frac{N-K(i-1)}{N-K(i)}\hspace{1 mm} v(i-1) +
\frac{1}{N-K(i)}\hspace{1 mm} d_{l}(i-1)^2
\end{equation}

where

\begin{equation}
\label{d} d_{l}(i-1)^2 = \frac{n_{l}(i-1)n_{l+1}(i-1)}{n_{l}(i-1) +
n_{l+1}(i-1)}\hspace{1 mm}[a_{l}(i-1)-a_{l+1}(i-1)]^2
\end{equation}

Similarly, we obtain

\begin{equation}\label{ww}
w(i)= \frac {K(i-1)-1}{K(i)-1} \hspace{1 mm} w(i-1)
-\frac{1}{K(i)-1} \hspace{1 mm} d_{l}(i-1)^2
\end{equation}

The equations \ref{vv}, \ref{ww} give a recursive formulation to
calculate the within and between class variances. We see that the
only term controlling the increment of $v(i)$ and the decrement of
$w(i)$ is the term $d_{l}(i-1)^2$, and thus minimizing
$d_{l}(i-1)^2$ helps us in optimizing $q(i)$.

\subsection{Algorithmic Implementation}

As above, let $M$ be the number of classes desired after
segmentation. The stepwise implementation of the above algorithm is
given below:
\begin{itemize}
\item Initialize the number of classes by the number of gray-levels ($K(0)=G$).
\end{itemize}

\begin{itemize}
\item Calculate $d_{k}^2$ for all pairs of adjacent classes for the $0^{th}$ stage.
\end{itemize}

\begin{itemize}
\item For each stage $i\leftarrow i+1$, until $K(i) \geq M$, Do:
\end{itemize}

\begin{enumerate}
\item Combine the $l^{th}$ and $(l+1)^{th}$ segments, where $l$ is the
index for which $d_{l}^2$ yields its minimum value.

\item Update the necessary as follows:
\begin{eqnarray*}
K(i) &=& K(i-1)-1 \\ a_{l}(i) &=& \frac{n_{l}(i-1)a_{l}(i-1) +
n_{l+1}(i-1)a_{l+1}(i-1)}{n_{l}(i-1)+n_{l+1}(i-1)} \\ n_{l}(i) &=&
n_{l}(i-1)+n_{l+1}(i-1) \\ a_{l-m}(i) &=& a_{l-m}(i-1); \hspace{1.12
in}m = 1, 2, \ldots l-1 \\ a_{l+m}(i) &=& a_{l+m+1}(i-1); \hspace{1
in}m = 1, 2, \ldots K(i)-l \\ n_{l-m}(i) &=& n_{l-m}(i-1);
\hspace{1.12 in}m = 1, 2, \ldots l-1 \\ n_{l+m}(i) &=&
n_{l+m+1}(i-1); \hspace{1 in}m = 1, 2, \ldots K(i)-l \\ d_{l-1}(i)^2
&=& \frac{n_{l-1}(i)n_{l}(i)}{n_{l-1}(i) + n_{l}(i)}\hspace{1
mm}[a_{l-1}(i)-a_{l}(i)]^2 \\ d_{l}(i)^2 &=&
\frac{n_{l}(i)n_{l+1}(i)}{n_{l}(i) + n_{l+1}(i)}\hspace{1
mm}[a_{l}(i)-a_{l+1}(i)]^2 \\ d_{l-1-m}(i)^2 &=& d_{l-1-m}(i-1)^2;
\hspace{1 in}m = 1, 2, \ldots l-2 \\ d_{l+m}(i)^2 &=&
d_{l+1+m}(i-1)^2; \hspace{1 in}m = 1, 2, \ldots K(i)-l-1
\end{eqnarray*}
\item The cumulative sums of the $n_{k}$'s yield us the values of thresholds.
\end{enumerate}

\section{Multilevel thresholding and Computational Advantage over Otsu}
The algorithm proposed can be used for multilevel image thresholding
as well as binarization. For having a particular number of classes,
we have to stop at a stage when the number of levels equals the
number of classes desired. This method has the advantage of being
computationally much more efficient than Otsu's, in spite of having
a very similar criterion function. We also know that of all methods
possible, Otsu's technique will have the minimum
Peak-Signal-to-Noise-Ratio (PSNR) value because PSNR acts like the
criterion function for Otsu's algorithm. We theoretically evaluate
the performance of the proposed algorithm and compare it with
conventional Otsu's method for multilevel thresholding. For
segmenting an image with $G$ gray-levels using $n$ thresholds,
Otsu's exhaustive search method searches $G\choose n$ different
combinations of thresholds, which can be approximated to $G^{n}$ for
$n<<G$. Thus the time complexity is exponential in the number of
thresholds.

\emph{\textbf{Claim}}: The proposed method can calculate all the possible thresholds in the multilevel case in $O(G^{2})$ time.

\textbf{\emph{Proof}}: We can divide the algorithm into three parts:
\begin{enumerate}
\item Finding the minima of $d_{l}^2$, which takes $O(K(i))$ steps for the $i^{th}$ iteration.
\item Updating the parameters, which can be done in $O(1)$ time.
\item Writing down the thresholds will take at most $O(K(i))$ time. We can also store the values efficiently by storing only the classes being merged at each stage.
\end{enumerate}
Thus, if we run the entire algorithm until we are left with two
classes ($G-1$ iterations), it will take $O(G^{2})$ time, thus
providing a tremendous edge in terms of computational time over the
exponentially complex Otsu's thresholding. We also point out here
that most of the hierarchical thresholding algorithms (for
example~\cite{Arifin}) will also consume $O(G^{2})$ time, but the
proposed method, because of its unique recursive formulation, incurs
low costs even in terms of the actual number of computations.

\section{Results and Observations}
\begin{figure}[!hbtp]
\centering
    \subfigure[Bacteria]{\includegraphics[width=25mm]{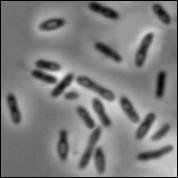}}
    \subfigure[Coins]{\includegraphics[width=25mm]{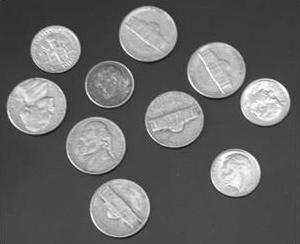}}
    \subfigure[Fish]{\includegraphics[width=25mm]{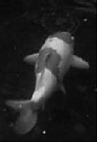}}
    \subfigure[Things]{\includegraphics[width=25mm]{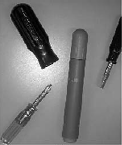}}
    \subfigure[Rice]{\includegraphics[width=25mm]{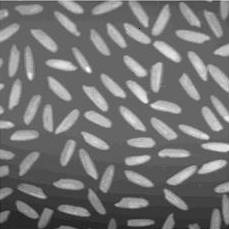}}
    \caption{Various Images we use for comparison.}
    \label{figuree1}
\end{figure}

\begin{figure}[!hbtp]
\centering
    \subfigure[Ground]{\includegraphics[width=25mm]{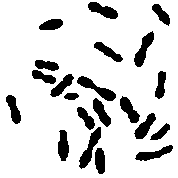}}
    \subfigure[Arifin]{\includegraphics[width=25mm]{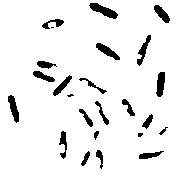}}
    \subfigure[Otsu]{\includegraphics[width=25mm]{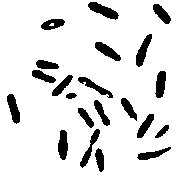}}
    \subfigure[KI]{\includegraphics[width=25mm]{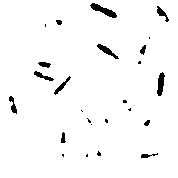}}
    \subfigure[Kwon]{\includegraphics[width=25mm]{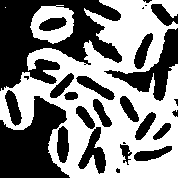}}
    \subfigure[Proposed]{\includegraphics[width=25mm]{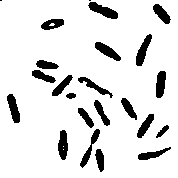}}
    \caption{Bacteria Image results for various methods}
  \label{bacteria}
\end{figure}

\begin{figure}[!htp]
\centering
    \subfigure[Ground]{\includegraphics[width=25mm]{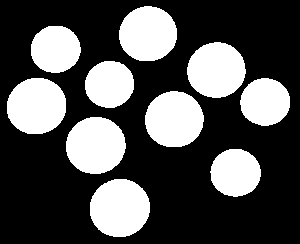}}
    \subfigure[Arifin]{\includegraphics[width=25mm]{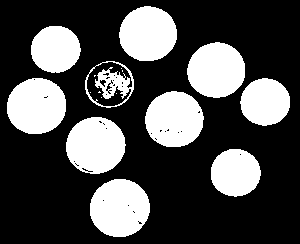}}
    \subfigure[Otsu]{\includegraphics[width=25mm]{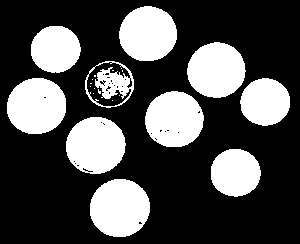}}
    \subfigure[KI]{\includegraphics[width=25mm]{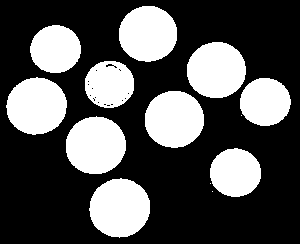}}
    \subfigure[Kwon]{\includegraphics[width=25mm]{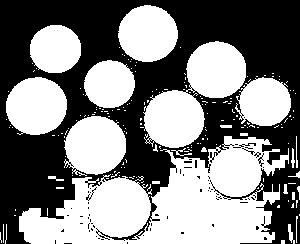}}
    \subfigure[Proposed]{\includegraphics[width=25mm]{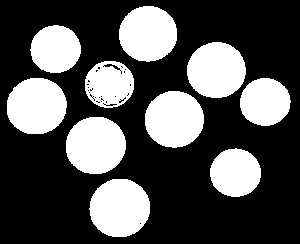}}
    \caption{Coins Image results for various methods}
  \label{coins}
\end{figure}

\begin{figure}[!htp]
\centering
    \subfigure[Ground]{\includegraphics[width=25mm]{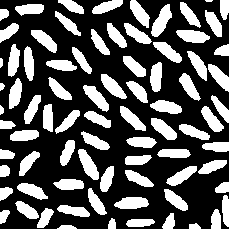}}
    \subfigure[Arifin]{\includegraphics[width=25mm]{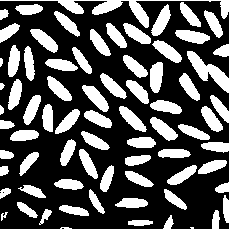}}
    \subfigure[Otsu]{\includegraphics[width=25mm]{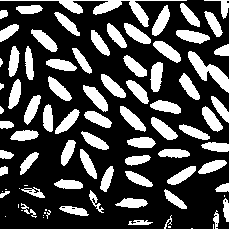}}
    \subfigure[KI]{\includegraphics[width=25mm]{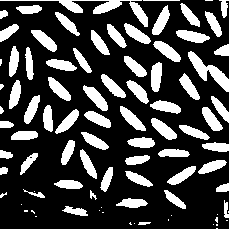}}
    \subfigure[Kwon]{\includegraphics[width=25mm]{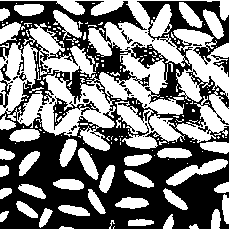}}
    \subfigure[Proposed]{\includegraphics[width=25mm]{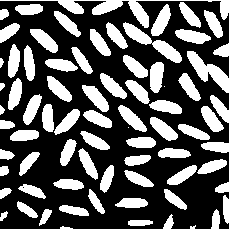}}
    \caption{Rice Image results for various methods}
  \label{fig:rice}
\end{figure}

\begin{figure}[!hbtp]
\centering
    \subfigure[Ground]{\includegraphics[width=25mm]{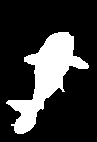}}
    \subfigure[Arifin]{\includegraphics[width=25mm]{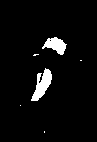}}
    \subfigure[Otsu]{\includegraphics[width=25mm]{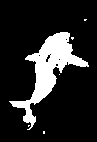}}
    \subfigure[KI]{\includegraphics[width=25mm]{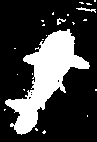}}
    \subfigure[Kwon]{\includegraphics[width=25mm]{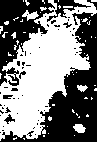}}
    \subfigure[Proposed]{\includegraphics[width=25mm]{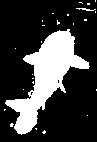}}
    \caption{Fish Image results for various methods}
  \label{fish}
\end{figure}

\begin{figure}[!hbtp]
\centering
    \subfigure[Ground]{\includegraphics[width=25mm]{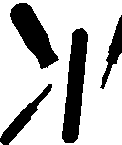}}
    \subfigure[Arifin]{\includegraphics[width=25mm]{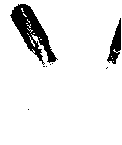}}
    \subfigure[Otsu]{\includegraphics[width=25mm]{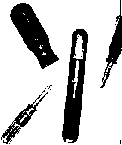}}
    \subfigure[KI]{\includegraphics[width=25mm]{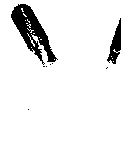}}
    \subfigure[Kwon]{\includegraphics[width=25mm]{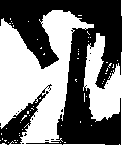}}
    \subfigure[Proposed]{\includegraphics[width=25mm]{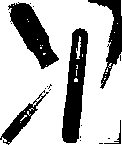}}
    \caption{Things Image results for various methods}
  \label{things}
\end{figure}

For evaluating the performance of the proposed algorithm, we have
implemented the method on a wide variety of images for both
binarization and multilevel thresholding. The image set we used for
comparing the performance of our algorithm with other methods for
the bilevel case is shown in Fig. \ref{figuree1}. Along with the
results obtained using the proposed technique, those of four other
methods have been shown - Otsu's~\cite{Otsu}, Kwon's~\cite{kwon},
Kittler-Illingworth's~\cite{Kittler} and
Arifin-Asano's~\cite{Arifin}.
The results are compared with the ground truth images generated by
manual thresholding (See Figs. \ref{bacteria}, \ref{coins},
\ref{fig:rice}, \ref{fish}, \ref{things})  We observe that our
method works at least at par with the other methods, and in most of
the cases produces a superior classification.

In order to compare the quality of the thresholded images
quantitatively, we have evaluated the performance of the proposed
method against the four different methods, using the criteria of
misclassification error (ME) and relative foreground area error
(RAE)~\cite{Sezgin}.

ME is defined in terms of the correlation of the images
with human observation. It corresponds to the ratio of background
pixels wrongly assigned to foreground, and vice versa.
ME can be simply expressed as:

\begin{equation}
ME= 1-\frac {|B_{O}\bigcap B_{T}|+|F_{O}\bigcap F_{T}|}{B_{O}+F_{O}}
\end{equation}

where background and foreground are denoted by $B_{O}$ and $F_{O}$
for the original image (the ground truth image), and by $B_{T}$ and
$F_{T}$ for the test image, respectively.

RAE measures the number of discrepancies in the thresholded image
with respect to the reference image, taking the area of the
foreground into account. It is defined as

\begin{equation}
RAE=\left\{\begin{array}{cl}
    \frac {A_{O}-A_{T}}{A_{O}}, &  A_{O}>A_{T} \\
    \frac {A_{T}-A_{O}}{A_{T}}, & A_{T}\geq A_{O}
       \end{array}\right.
\end{equation}
where $A_{O}$ is the area of the reference image, and $A_{T}$ is the
area of thresholded image.

In Table~\ref{table1} we show the ME and RAE for the various images
under consideration.

\begin{table}[t]
\centering
\begin{tabular}{cccccc}
    \hline
Image &  Arifin   &   Otsu  & KI & Kwon   & Proposed  \\
    \hline
$ME$ & & & & & \\
Bacteria & 0.0857 & 0.0432 & 0.1253 & 0.3316 & 0.0459 \\
Rice & 0.0589 & 0.0741 & 0.0924 & 0.1719 & 0.0609 \\
Fish & 0.1745 & 0.0673 & 0.0356 & 0.2860 & 0.0263 \\
Coins &  0.0163 &  0.0135 & 0.0076 & 0.1887 & 0.0066 \\
Things & 0.1759 & 0.0432 & 0.1772 & 0.2669 & 0.0515 \\
$RAE$ & & & & & \\
Bacteria & 0.4629 & 0.1893 & 0.7027 & 0.6394 & 0.2200 \\
Rice & 0.0318 & 0.1495 & 0.2311 & 0.2973 & 0.0325 \\
Fish & 0.7981 & 0.2929 & 0.1276 & 0.5675 & 0.0859 \\
Coins & 0.0514 & 0.0426 & 0.0200 & 0.3726 & 0.0095 \\
Things & 0.1912 & 0.0448 & 0.1923 & 0.3356 & 0.0204 \\

    \hline
\end{tabular}
\caption{Comparison of ME and RAE for various images on the methods}
\label{table1}
\end{table}

In this section, we also consider another measure called the Peak
Signal-to-Noise Ratio (PSNR) of the segmented image. Let $S$ be the
source image and $T$ be the image obtained after thresholding ($S$
means $S(i,j)$ and $T$ means $T(i,j)$). $N$ is the total number of
pixels in the image. Let us assume for the time being that the
number of gray-levels is 256. Then, the mean square error (MSE) and
the PSNR are given by:

\begin{eqnarray}
\label{MSE} MSE &=& \frac{\sum{[S - T]^2}}{N} \\
\label{PSNR} PSNR &=& 10\log_{10}{\left(\frac{255^2}{MSE}\right)},
\end{eqnarray}

where the summation in MSE is over all the pixels in the image and
the PSNR value obtained above is in dB (decibels).

PSNR is used to determine the overall quality and error of
thresholded image. Several test images were taken and PSNR was
calculated for thresholded images for multiple levels using the
proposed technique. The results are shown in Table~\ref{TablePSNR}.
We can see that the value of PSNR grows in general with the number
of classes, and this combined with the computational advantage
discussed above makes the process valuable. The table of PSNR values
gives us an idea that the method can be used for ``fast" lossy
compression in real time in a general setting owing to high PSNR
values. We give examples of natural and smooth pictures in addition
to the images considered above because the objective here is
compression and not object recognition.

\begin{table}[t]
\centering
\begin{tabular}{cccccc}
    \hline
Image &  2 level   &   3 level  & 5 level & 10 level   & 25 level  \\
    \hline
Lena &  20.3 &  23.6 & 27.0 & 33.4 & 41.1 \\
Baboon & 20.1 & 24.1 & 27.8 & 32.8 & 40.6 \\
Peppers & 19.2 & 21.6 & 26.4 & 32.1 & 39.8 \\
Rice & 21.4 & 23.6 & 28.5 & 34.1 & 41.8 \\
Fish & 21.3 & 26.6 & 30.8 & 36.2 & 43.4\\
Things & 18.7 & 21.2 & 26.7 & 31.6 & 39.5 \\

    \hline
\end{tabular}
\caption{PSNR Values for various levels of image thresholding by proposed method}
\label{TablePSNR}
\end{table}

\begin{figure}  [!hbtp]
\centering
    \subfigure[Gray]{\includegraphics[width=25mm]{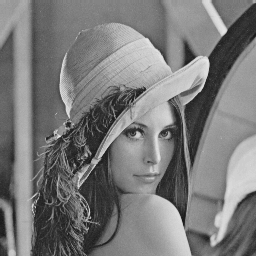}}
    \subfigure[2 Level]{\includegraphics[width=25mm]{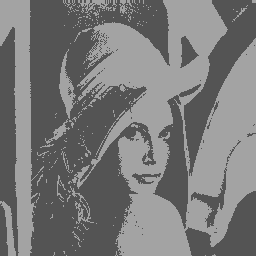}}
    \subfigure[3 Level]{\includegraphics[width=25mm]{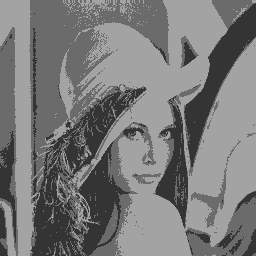}}
    \subfigure[5 Level]{\includegraphics[width=25mm]{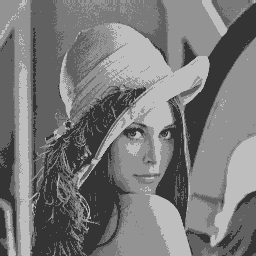}}
    \subfigure[10 Level]{\includegraphics[width=25mm]{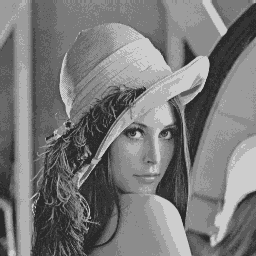}}
    \subfigure[25 Level]{\includegraphics[width=25mm]{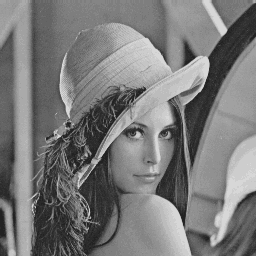}}
    \caption{Lena image thresholding Image results for various levels.}
    \label{psnrimage}
\end{figure}

\begin{figure}[!hbtp]
\centering
    \subfigure[Gray]{\includegraphics[width=25mm]{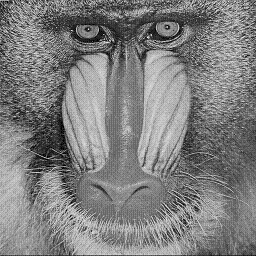}}
    \subfigure[2 Level]{\includegraphics[width=25mm]{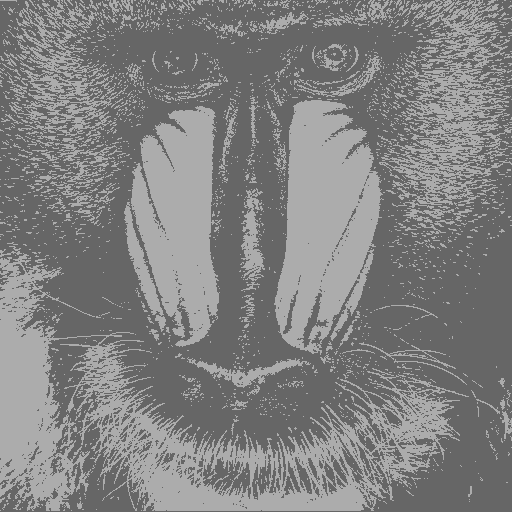}}
    \subfigure[3 Level]{\includegraphics[width=25mm]{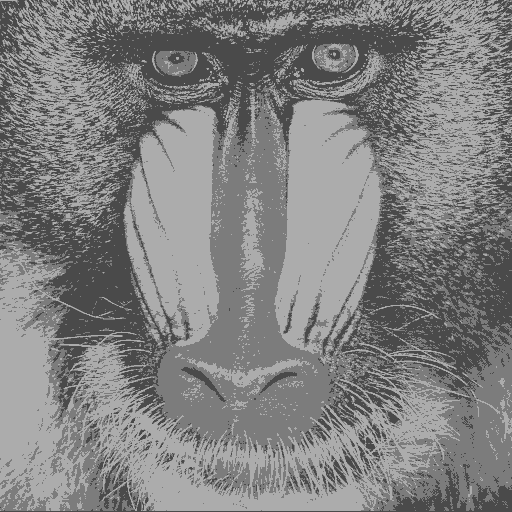}}
    \subfigure[5 Level]{\includegraphics[width=25mm]{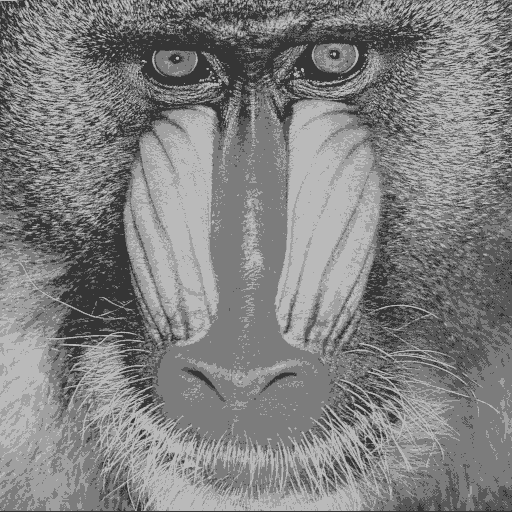}}
    \subfigure[10 Level]{\includegraphics[width=25mm]{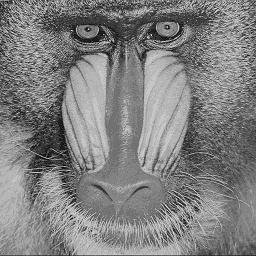}}
    \subfigure[25 Level]{\includegraphics[width=25mm]{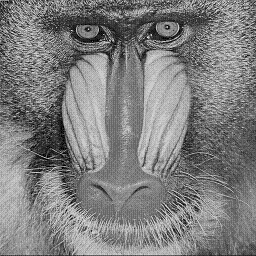}}
    \caption{Baboon image thresholding Image results for various levels.}
    \label{psnrimage1}
\end{figure}

\begin{figure}[!hbtp]
\centering
    \subfigure[Gray]{\includegraphics[width=25mm]{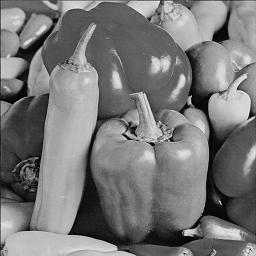}}
    \subfigure[2 Level]{\includegraphics[width=25mm]{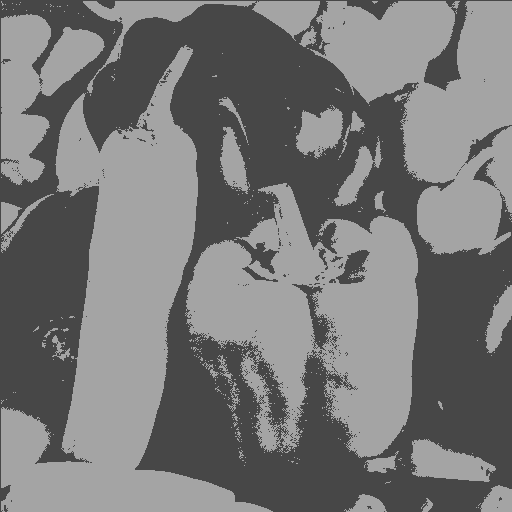}}
    \subfigure[3 Level]{\includegraphics[width=25mm]{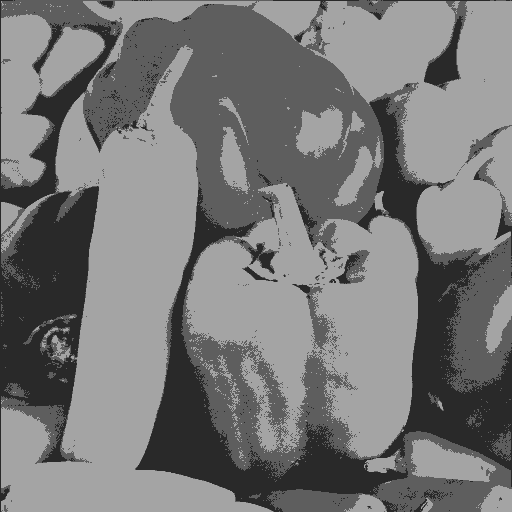}}
    \subfigure[5 Level]{\includegraphics[width=25mm]{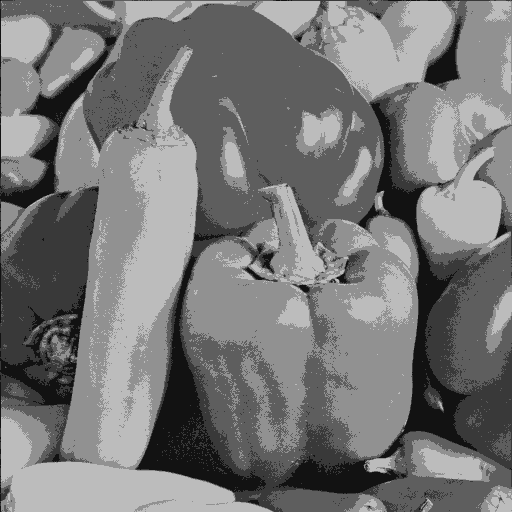}}
    \subfigure[10 Level]{\includegraphics[width=25mm]{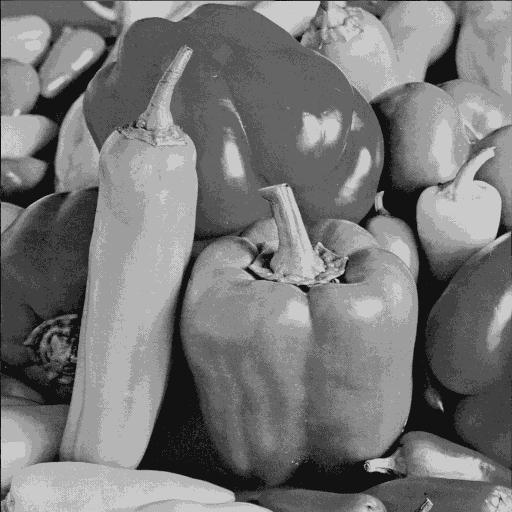}}
    \subfigure[25 Level]{\includegraphics[width=25mm]{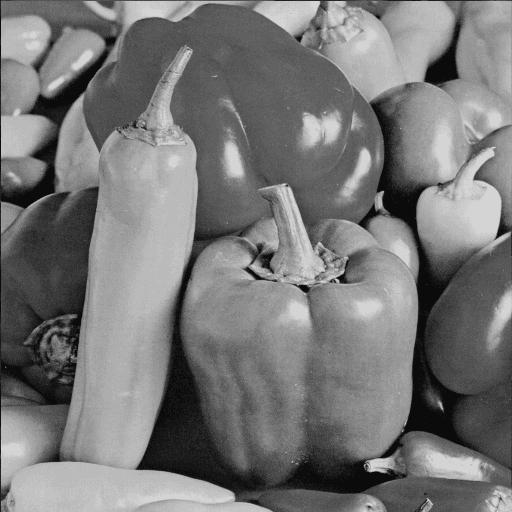}}
    \caption{Peppers image thresholding Image results for various levels.}
     \label{psnrimage2}
\end{figure}

We show the original and thresholded images for some well known
images in Figs. \ref{psnrimage}, \ref{psnrimage1}, \ref{psnrimage2}
obtained using the proposed method. Note that for the bilevel case,
we show images with gray-levels equal to the means of the two
classes.

\section{Conclusion}
We have proposed a novel image segmentation technique based on
unbiased variance estimators. The proposed method is compared with
other global histogram-based thresholding methods. The various
results show the efficiency of the proposed method with respect to
object recognition and also for compression owing to a high PSNR
associated with it. The method is also computationally very
advantageous because of its polynomial running time.

\section*{Acknowledgment}

We would like to thank Dr. Denis Heliot of Schlumberger Technology
Corporation, under whose supervision a part of the work began. We
also thank Prof. Somnath Sengupta of IIT Kharagpur for his valuable
suggestions.

\ifCLASSOPTIONcaptionsoff
  \newpage
\fi


\begin{thebibliography}{1}

\bibitem{chietal}
Z.~Chi, H.~Yan, T.~Pham, ``Fuzzy Algorithms: With applications to
images processing and pattern recognition,'' World Scientific, Singapore, 1996..
\bibitem{Sezgin}
M.~Sezgin and B.~Sankur, ``Survey over image thresholding
techniques and quantitative performance evaluation'', \emph{Journal of
Electronic Imaging}, 13(1), 146-165, 2004.
\bibitem{Niblack}
W.~Niblack, ``An Introduction to Digital Image Processing,''
Englewood Cliffs, N. J.: Prentice Hall, 115-116, 1986.
\bibitem{Otsu}
N. Otsu, ``A threshold selection using gray level histograms,''
\emph{IEEE Trans. Systems Man Cybernetics}, 9, 62-69, 1979.
\bibitem{abutaleb}
A.~S.~Abutaleb, ``Automatic thresholding of gray level pictures
using two-dimensional entropy,'' Comput. Vision Graphics Image
Process. 47, 22-32, 1989.
\bibitem{kwon}
S.~H.~Kwon, ``Threshold selection based on cluster analysis,'' \emph{Pattern
Recognition Letters}, 25, 1045–1050, 2004.
\bibitem{Reddi}
S.~S.~Reddi, S.~F.~Rudin, H.~R.~Keshavan, ``An Optical
Multiple Threshold Scheme for Image Segmentation'', \emph{IEEE Trans.
System Man and Cybernetics}, 14, 661-665, 1984.
\bibitem{Ridler}
T.~W.~Ridler, S.~Calward, ``Picture Thresholding Using an
Iterative Selection Method'', \emph{IEEE Trans. Systems, Man and
Cybernetics}, 8, 630-632, 1978.
\bibitem{Arora}
S.~Arora, J.~Acharya, A.~Verma, P.~K.~Panigrahi, ``Multilevel Thresholding for Image Segmentation through a Fast Statistical Recursive Algorithm,'' \emph{Pattern Recognition Letters}, 29, 119–125, 2008.
\bibitem{Sahoo}
P.~K.~Sahoo, S.~Soltani, A.~K.~C.~Wong, ``SURVEY: A survey of
thresholding techniques'', \emph{Comput. Vision Graphics Image
Process}, 41, 233-260, 1988.
\bibitem{Arifin}
A.~Z.~Arifin, A.~Asano, ``Image segmentation by histogram
thresholding using hierarchical cluster analysis,'' \emph{Pattern
Recognition Letters}, 27, 1515-1521, 2006.
\bibitem{Kittler}
 J.~Kittler, J.~Illingworth, ``Minimum error thresholding,'' \emph{Pattern Recognition}, 19(1), 41-47, 1986.
\bibitem{Papa}
N.~Papamarkos, B.~Gatos, ``A new approach for multilevel
threshold selection,'' Graphics Models Image Process, 56, 357-370, 1996.
\bibitem{Moghaddamjoo}
A.~Moghaddamjoo, ``Optimum Well-Log Signal Segmentation,''\emph{IEEE
Transactions on Geoscience and Remote Sensing}, 27(5), 633 - 641, 1989.
\bibitem{Acharya}
J.~Acharya, G.~Sreechakra, J.~C.~Goswami, D.~Heliot, ``Hierarchical
Zonation Technique to Extract Common Boundaries of a Layered Earth
Model,'' Proceedings of the \emph{IEEE Antenna and Propagation
Society Symposium}, Hawaii, 2007.
\end{thebibliography}
\end{document}